\documentclass[10pt,a4paper,twoside]{article}
\usepackage[colorlinks,urlcolor=blue,linkcolor=blue,citecolor=blue]{hyperref}

\usepackage{color,array}
\usepackage{amsfonts}
\usepackage{algorithmic}
\usepackage{algorithm}
\usepackage{array}
\usepackage{textcomp}
\usepackage{stfloats}
\usepackage{url}
\usepackage{verbatim}
\usepackage{dirtytalk}
\usepackage{xcolor}
\usepackage{bbm}
\usepackage{graphicx}
\usepackage{cite}
\input{spp.dat}

\begin{document}

%--------------------------------------------------
%  Fill in the paper's title in Sentence case
%  Titles beginning with articles (A, An, The) are discouraged
%--------------------------------------------------
\title{\TitleFont World Models Unlock Optimal Foraging Strategies in Reinforcement Learning Agents}

%--------------------------------------------------
%  Authors with different affiliations
%--------------------------------------------------
% Manuel: Biomedica (1) + Bancolombia (2) + Corresponsal (*)
\author[2]{Yesid Fonseca\authorsep} 
\author[1,2]{Manuel S. Ríos\authorsep} 
\author[3]{Nicanor Quijano\authorsep}
% Luis F. Giraldo: Biomedica (1)
\author[1]{Luis F. Giraldo\lastauthorsep} 

% Affiliations List
\affil[1]{Department of Biomedical Engineering, Universidad de los Andes, Bogotá, Colombia}
\affil[2]{Center of Excellence in Analytics, Artificial Intelligence, and Information Governance, Bancolombia, Colombia}
\affil[3]{Department of Electrical and Electronics Engineering, Universidad de los Andes, Bogotá, Colombia}

% Correspondence Email
\affil[*]{\corremail{y.fonseca@uniandes.edu.co}}

%--------------------------------------------------
%  For three or more authors with the same affiliation please use this block
%--------------------------------------------------

% \author[*]{Author M.~Surname\authorsep}
% \author[ ]{Bauthor D.~Surname~Jr.\authorsep}
% \author[ ]{Cauthor D.~Surname~III\lastauthorsep}
% \affil[ ]{Department of Science, XXX University, Country}
% \affil[*]{\corremail{amsurname@university.edu} }

%--------------------------------------------------
%  For authors with different affiliations please use the following block
%--------------------------------------------------
% \author[1*]{Author M.~Surname\authorsep}
% \author[2]{Bauthor D.~Surname~Jr.\authorsep}
% \author[1,2]{Coauthor G.~Surname~III\authorsep}
% % !!! Please take note that the last author separation is \lastauthorsep instead of \authorsep
% \author[3]{Dauthor G.~Surname\lastauthorsep}
% \affil[1]{Department of Physics, DD University, Country}
% \affil[2]{Department of Science, XX University, Country}
% \affil[3]{Physics Institute, Country}
% \affil[*]{\corremail{amsurname@university.edu} }

\begin{abstract}
\noindent
%--------------------------------------------------
% Include abstract and keywords here
%--------------------------------------------------
Patch foraging involves the deliberate and planned process of determining the optimal time to depart from a resource-rich region and investigate potentially more beneficial alternatives. The Marginal Value Theorem (MVT) is frequently used to characterize this process, offering an optimality model for such foraging behaviors. Although this model has been widely used to make predictions in behavioral ecology, discovering the computational mechanisms that facilitate the emergence of optimal patch-foraging decisions in biological foragers remains under investigation. Here, we show that artificial foragers equipped with learned world models naturally converge to MVT-aligned strategies. Using a model-based reinforcement learning agent that acquires a parsimonious predictive representation of its environment, we demonstrate that anticipatory capabilities, rather than reward maximization alone, drive efficient patch-leaving behavior. Compared with standard model-free RL agents, these model-based agents exhibit decision patterns similar to many of their biological counterparts, suggesting that predictive world models can serve as a foundation for more explainable and biologically grounded decision-making in AI systems. Overall, our findings highlight the value of ecological optimality principles for advancing interpretable and adaptive AI.


\end{abstract}

\maketitle
\thispagestyle{empty}  
\pagestyle{empty}

%--------------------------------------------------
% the main text of your paper begins here
%--------------------------------------------------
\section{Introduction}
Optimality in decision-making is a hallmark of animal behavior, particularly evident in essential survival tasks such as foraging \cite{CHARNOV1976129,Krebs1994,STEPHENS1986}. The imperative nature of these tasks has led to the evolution of behaviors that maximize resource acquisition while minimizing effort and risk. Foraging theory within behavioral ecology provides a framework to study these behaviors, encompassing patch-foraging models and predator–prey dynamics \cite{Macarthur1966, COWIE1977, Kacelnik1984}. These models show how agents adapt to environmental cues, informing their resource exploitation decisions and search for new patches. According to optimal foraging theory, in environments where resources are distributed in patches, optimized decisions made by an agent can be described by the Marginal Value Theorem (MVT) \cite{CHARNOV1976129}. The MVT is an optimality model that considers elements such as travel costs between patches and energy expenditure within patches to predict an agent’s behavior. Although some discrepancies have been observed between MVT predictions and the decisions of real foraging organisms \cite{NEURIPS2020_da97f65b, Constantino2015,Overharvesting2023}, the MVT remains a valuable tool for understanding adaptive foraging behavior.

Given the broad use of this optimality model, an important question arises: what mechanisms enable learning agents to exhibit the optimal behaviors predicted by the MVT? Although the MVT specifies optimal outcomes, the learning and decision-making mechanisms that give rise to such behaviors remain incompletely understood \cite{MILLER201777}. Computational models based on reinforcement learning (RL) have been proposed as a way to shed light on this question. RL enables us to study foragers as learning agents that engage in sequential decision-making and seek to optimize their actions based on reward \cite{NEURIPS2020_da97f65b, wispinski2023adaptive}. Most prior studies rely on model-free RL algorithms, where agents learn directly from past experience without forming an internal representation of the environment. While such analyses reveal similarities between biological and artificial agents, the mechanisms behind these behaviors remain unclear. In this work, we implement a patch-foraging scenario to evaluate RL algorithms, with particular emphasis on agents that learn world models.

Research suggests that, to effectively process the immense volume of information encountered daily, the brain constructs a predictive understanding of the spatial and temporal structure of the environment \cite{Clark_2013, Quiroga2005, Friston2010}. Under evolutionary pressures, animals develop internal world models based on their limited sensory input, and these models guide their decisions and actions. We posit that similar internal representations may enable artificial agents to make decisions consistent with the MVT. By maintaining an internal model of the world, our artificial forager can anticipate future events and act accordingly. From an AI perspective, understanding how such internal models give rise to adaptive behaviors offers a principled framework for studying explainable decision-making mechanisms beyond classical reward-driven optimization.

In this study, we investigate how model-based agents, specifically those that learn compact world models capable of anticipating future environmental changes, behave in classic patch-foraging tasks. Rather than proposing a new algorithm, our contribution is conceptual: we demonstrate that when an agent learns to predict the consequences of its actions, its decisions naturally align with the optimal patch-leaving strategies described by the MVT. Through controlled simulations and comparisons with a model-free baseline, we show that anticipatory representations support more adaptive and ecologically coherent behavior, offering insights into the computational principles that may underlie optimality in both artificial and biological foragers.

Next, we present the formulation of the Marginal Value Theorem, the reinforcement learning model, and the adapted environment used to evaluate the agent’s patch-foraging capabilities (Section II). We then show, through simulations, how agents with anticipatory capabilities exhibit behaviors consistent with the MVT (Section III), followed by a discussion analyzing the role of world models in the emergence of optimality and connecting our results with observations in biological foragers (Section IV). Finally, we conclude the paper with remarks and directions for future work (Section V).
\section{Background}
\subsection{MVT Formulation}
To facilitate a comparative analysis between MVT and RL algorithms, which share parallels between their respective decision-making capabilities, it is crucial to grasp the formality of MVT. This enables a seamless transition from the variables analyzed by the MVT model to those measurable on a computational scenario of patch foraging, which is the specific context in which we work and present our contribution.

Consider a predator in a fragmented habitat, i.e., the food is distributed across numerous lots or patches \cite{CHARNOV1976129}. In  these types of habitats, predators encounter resources unevenly distributed across patches, necessitating strategic foraging decisions. Each patch offers food, but with the caveat of diminishing returns: the longer a predator remains, the less food it efficiently gathers. This dynamic is captured by the function $ h_i(T) $, denoting food acquired in time $ T $ within a patch type $ i $. This function follows an asymptotic growth pattern indicating reduced foraging efficiency over time. Crucially, the predator faces a trade-off between staying in a patch and moving to another, considering the energy and time costs of travel. This posits a repetitive environment, where the predator's objective is to optimize its net rate of energy intake. This involves balancing the decreasing food gains in a current patch against the potential benefits of exploiting a new patch, thereby maximizing energy attainment with minimal effort.

Taking into account the aforementioned context, we introduce the key variables essential to the development of the MVT. Let $P_i$ denote the proportion of patches of type $i$. Regarding energy expenditure, $E_T$ represents the cost per unit time for inter-patch movement, while $E_{si}$ corresponds to the cost per unit time for intra-patch search of type $i$. We define $h_i(T)$ as the energy assimilated after $T$ units of time in patch $i$. Consequently, the net assimilated energy $g_i(T)$ is defined as the difference between gained energy and search costs, given by the equation $g_i(T) = h_i(T) - E_{si}T$.
% \begin{align*}
%     & P_i  : \text{Proportion of patches of type } i.\\
%     & E_T  : \text{Energy cost per unit time for inter-patch movement.}\\
%     & E_{si} : \text{Energy cost per unit time for intra-patch search of type } i.\\
%     & h_i(T) : \text{Energy assimilated after } T \text{ units of time in patch } i.\\
%     & g_i(T) : \text{Net assimilated energy, given by } g_i(T) = h_i(T) - E_{si}T.
% \end{align*}
% \begin{align*}
%     P_i    &: \text{Proportion of patches of type } i.\\
%     E_T    &: \text{Energy cost per unit time for inter-patch movement.}\\
%     E_{si} &: \text{Energy cost per unit time for intra-patch search of}
%     \\
%            &\quad \text{type } i.\\
%     h_i(T) &: \text{Energy assimilated after } T \text{ units of time in patch } i.\\
%     g_i(T) &: \text{Net assimilated energy, given by } \\
%            &\quad g_i(T) = h_i(T) - E_{si}T.
% \end{align*}

The net rate of energy intake, $E_n$, is defined as:
\begin{equation}
    E_n = \frac{\sum P_i g_i(T_i) - tE_T}{t + \sum P_iT_i},
    \label{eq:energynetrate}
\end{equation}
where $t$ is the travel time between patches. A crucial assumption is that the predator independently determines the duration of stay within a patch, unrelated to the time spent traveling between them. Given this independence, we optimize the time spent in patch $\left(T_j\right)$, for an individual patch type $j$, as follows: 

\begin{equation*}
E_n = \frac{P_jg_j(T_j) + \sum_{i \neq j} P_i g_i(T_i) - tE_T}{P_jT_j+\sum_{i \neq j} P_iT_i + t}, 
\end{equation*}
meaning that the rate of energy per unit time spent is defined by energy gained patch $j$ subtracted by energy spent seeking all intermediary patches $i \neq j$. For a set of patches at optimal foraging times, the gradient of the energy intake function, should be zero. By setting $ \frac{\partial E_n(T_i)}{\partial T_i} = 0 $, we arrive at the theorem's prediction: $\frac{\partial g_j(T_j)}{\partial T_j} = E_n^*$.
%\begin{equation*}
 %  \frac{\partial g_j(T_j)}{\partial T_j} = E_n^* .
%\end{equation*}

This relationship solidifies the MVT as a quantitative model and enhances our comprehension of optimal foraging. Building on this, we explore algorithmic strategies that might emulate such efficiency in artificial systems. RL algorithms are well posed for exploration (under uncertainty), and are typically divided into model-free and model-based paradigms. Model-free RL learn from direct interactions, crafting value estimates based on accumulated experience and adjusting to prediction errors \cite{Sutton1991}. Conversely, model-based RL constructs a conceptual model of the environment, enabling the agent to forecast future states and rewards, thereby informing their strategic planning for longer-term horizons \cite{Kolling2017}. This predictive planning, intrinsic to model-based RL, echoes the anticipatory tactics found in natural foragers, equipping them to adeptly navigate the uncertain terrains of their ecosystems.

\subsection{Reinforcement Learning Model}
\begin{figure*}[ht!]
    \centering    
    \includegraphics[width=0.75\textwidth]{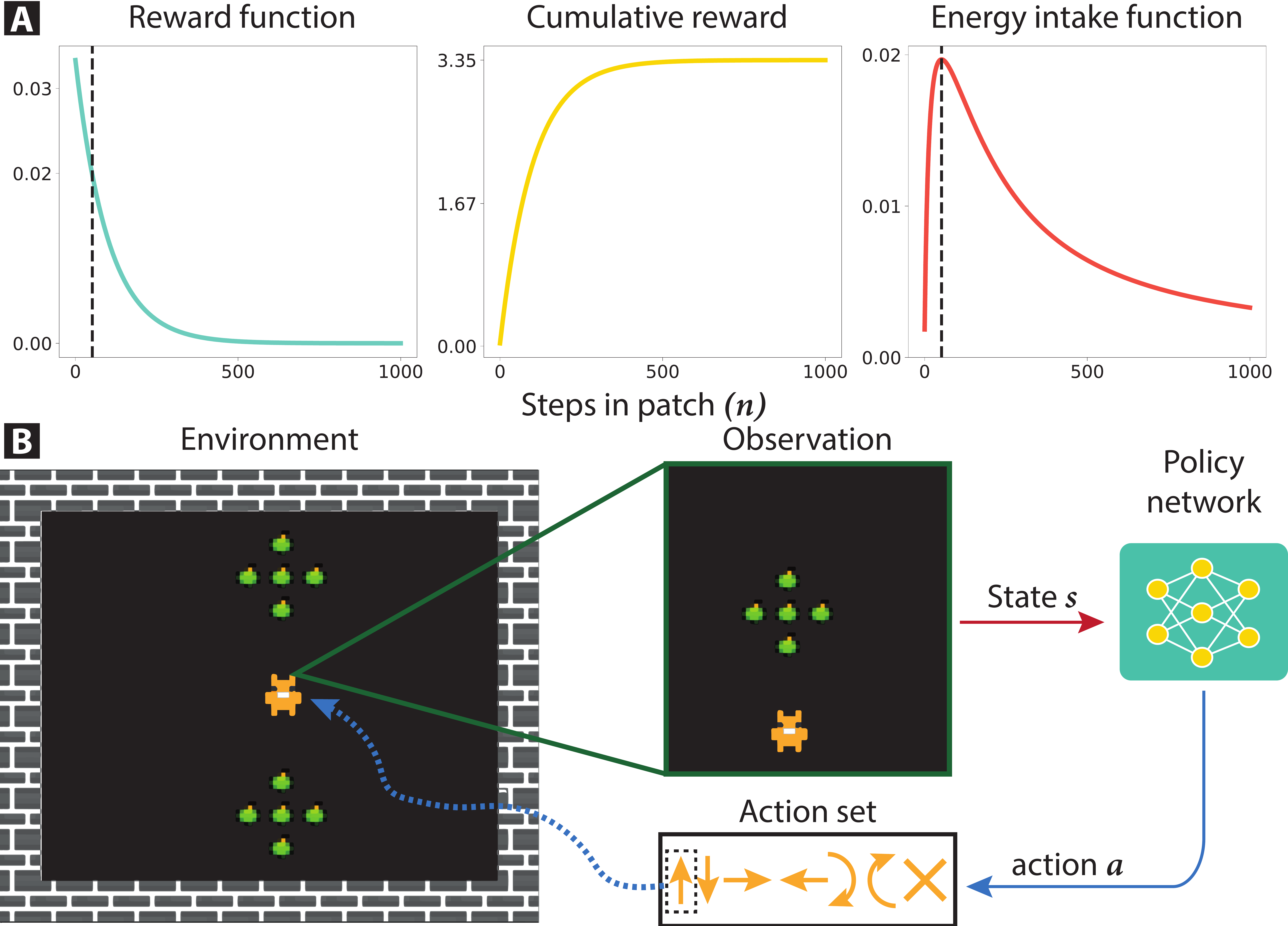}
    \caption{Representation of the patch foraging environment. (A) Exponentially decaying reward function $r(n)$, to the right is the cumulative sum of the reward function, and the graph at the far right is the function that is to be optimized through a model-based RL algorithm. The dashed line represents the value that maximizes the energy intake function $E_n(n)$, that is, the optimal number of steps to remain in a patch. (B) Interaction between the agent immersed in the melting pot environment adapted for the foraging task, the agent interacts with the environment through an action $a$ which generates a reward $r$ and a state $s$, which will go to a decision policy $\pi$.}
    \label{fig:EnvironmentDynamics}
\end{figure*}

In the patch foraging problem, the agent is immersed in a partially observable 2D game (Fig. \ref{fig:EnvironmentDynamics}B). In each state of the game, the agent takes actions based on a partial observation of the environment from a set of states and receives a reward. The agent must learn to interact with the environment in such a way that at the end of the game, the reward is maximized. We formulate this as follows: the agent is in an environment and is assigned a state \(s \in \mathcal{S}\) (where \(\mathcal{S}\) denotes a finite set of states), the agent has the ability to change to a new state \(s' \in \mathcal{S}\) through an action \(a \in \mathcal{A}\), and as a result of this transition, the agent receives a reward \(r \in \mathbb{R}\). This can be more tersely expressed by stating that the agent is attempting to solve a Partially Observable Markov Decision Process (POMDP). This means we have a tuple \(\mathcal{M} = (\mathcal{S}, \mathcal{A}, r, \mathcal{T}, \mathcal{O}, \gamma)\), where \((\mathcal{S}, \mathcal{A})\) represent the set of states and actions, \(r: \mathcal{S} \times \mathcal{A} \rightarrow \mathbb{R}\) is the reward function, \(\mathcal{T}: \mathcal{S} \times \mathcal{A} \rightarrow \Delta (\mathcal{S})\), is the stochastic transition function, where \(\Delta (\mathcal{S})\) denotes the set of discrete probability distributions over \(\mathcal{S}\), \(\mathcal{O}: \mathcal{S} \rightarrow \mathbb{R}^d \) is the observation function that specifies player $d$-dimensional view on the state space, and \(\gamma\) is a discount factor. The agent's goal is to find an behavioral policy \(\pi: \mathcal{O} \rightarrow \Delta (\mathcal{A})\) (written $\pi\left(a \mid o \right)$ based on its own observation \(o= O(s)\) and extrinsic reward \(r(s_t, a_t)\)) that maximizes the expected cumulative reward $V_{\pi}(s_0) = \mathbb{E}\left[ \sum_{t=0}^{\infty } \gamma^t r\left ( a_t, s_t \right ) \mid a_t \approx \pi_t , s_{t+1} \approx \mathcal{T} \left ( s_t,a_t \right) \right]
$. This is the adaptation for a single agent of what was deduced in \cite{Raphael2022}.

An agent represents a function \(\mathcal{Q}: \mathcal{O} \times \mathcal{A}\), this function is often a deep neural network known as \(\mathcal{Q}\)-Network, which seeks to map the observation of a state, \(o \in \mathcal{O}\), to the \textit{value} of taking action \(a \in \mathcal{A}\) in that state. The magnitude of this \textit{value} represents the expected return on taking that action, therefore, at time \(t\), the agent will prefer to take the action \(a_t = \arg \max_{a \in \mathcal{A}} \mathcal{Q}\)
\cite{sutton2018reinforcement}.

Within this reinforcement learning framework, agents are commonly categorized as model-free or model-based. Model-free agents update their policies purely through trial-and-error interaction with the environment, without learning an explicit representation of its dynamics. In contrast, model-based agents learn a compact predictive representation of how observations, states, and rewards evolve in response to actions, known as world model. A subset of these model-based architectures leverages the learned world model for imagination-based planning, internally simulating future trajectories before committing to an action in the real environment. In this work, we show that such predictive capabilities are particularly relevant in patch-foraging tasks, where anticipating future resource depletion is key to determining near-optimal patch-leaving decisions.

\subsection{Environment adaptation}

We have modified a single-agent reinforcement learning (SARL) environment in a two-dimensional world to simulate a scenario where an agent collects apples from a patch called \say{Commons Harvest} in the Melting Pot suite  \cite{leibo21a}. This setup mirrors a patch-foraging situation where resources are spread across two patches and offer exponentially decreasing rewards to the agent:
\begin{equation}
    r(n) =
    N\exp{\left(-\lambda n\right)} \mathbbm{1}_{\left \{ \text{in patch} \right \}},
    \label{eq:RewardFunction}
\end{equation}
where $n$ is the number of time steps that the agent takes inside the patch, and $N$ and $\lambda$ are constant parameters of the function. The rationale behind employing an exponentially decreasing reward structure in our SARL environment aligns with behavioral observations in animal studies. Intertemporal choice tasks traditionally contrast immediate against delayed rewards, interpreting a preference for immediacy as impulsivity. However, recent interpretations suggest that what appears as impulsivity may instead be an adaptive response to environmental variability and the inherent uncertainty of future rewards \cite{stevens2014intertemporal, Hayden2016}.

In our custom environment, the exponential decay of rewards reflects a realistic aspect of foraging, where the value of a resource diminishes as it is exploited. This decay mimics the diminishing returns animals might encounter in nature, thereby providing an agent with a scenario where the strategic allocation of time and effort becomes essential. The agent's interaction with this reward structure allows us to investigate its ability to develop a foraging strategy that balances the pursuit of immediate needs against the potential benefits of conserving resources for future use, a critical aspect of survival in natural ecosystems.

By incorporating this exponential reward function, we aim to replicate the decision-making dynamics inherent to patch foraging. The selection of the reward assigned to the agent $r(n)$ also aligns with mathematical properties outlined in \cite{STEPHENS1986}, as the $g(T)$ function should be well-defined, continuous, deterministic, and negatively accelerated. Thus, the gain function has the following characteristics:
\begin{equation*}
    g(T) :
    \begin{cases}
        g(0) = 0\\
        g'(0) > 0\\
        \exists \, \tilde{t} \mid g''(T) < 0, \, \forall t \geq \tilde{t},
    \end{cases}
\end{equation*}
where $g'(0)$ and $g''(T)$ represent the derivative of the function at time zero and the second derivative with respect to time. This implies that the net energy gain is zero when no time is spent foraging within a patch ($g(0) = 0$), that there is an initial phase where the energy gain increases ($g'(0) > 0$), and that the gain function ultimately displays a negatively accelerated trajectory over time, this means that there exists a specific threshold, denoted as $\tilde{t}$, beyond which the energy assimilable by an agent within a given patch gradually diminishes. This phenomenon arises due to the inherent finite nature of resources.

Note that Figure \ref{fig:EnvironmentDynamics}A shows a reward with exponential decay. 
However, when this reward is accumulated, it manifests as a function that fully satisfies the previously established conditions, i.e. a well-defined, continuous, deterministic, and negatively accelerated function. This indicates a correlation between $g(T)$ and $\sum_{i=1}^n r(i)$. Indeed, this relationship will serve as the foundation for calculating the marginal value rule within the RL framework. For this purpose, let us consider that the energy gain function is equivalent to the accumulated sum of rewards. Now, we must express the net rate of energy intake function in equation \eqref{eq:energynetrate} in terms of the reward. Assuming that there is only one type of patch, i.e.,  $P_i = 1$ and $E_T=0$, the function can be rewritten as:
\begin{equation}
E_n(n) = \frac{\sum_{i=1}^n r(i)}{\bar{x} + n},
\label{eq:EnergyIntakeAdapted}
\end{equation}
where $ \bar{x} $ represents the distance between patches. This assumption holds because if the agent's speed within Melting Pot is one square per step, then the time $ t $ can be expressed solely in terms of the distance between patches, as Charnov \cite{CHARNOV1976129} posits: \say{A simple assumption would have $ t $ proportional to the distance between patches divided by the predator’s speed of movement.} Another integral aspect of this $ E_n $ model is its ability to ascertain the optimal number of steps the agent should take within a patch $ n^*$, analogous to the step count that optimizes energy intake for a biological agent in a patch foraging setting.To formalize this condition, we propose to optimize equation \eqref{eq:EnergyIntakeAdapted}, determining the optimal number of steps that an agent should take within the patch, which can be computed as: $n^* = \arg \max_{n} E_n(n)$. This implies that there is a direct method for calculating the MVT within the 2D environment (Figure \ref{fig:EnvironmentDynamics}B) for the patch foraging task.

\section{Experiments}

\subsection{Architectures of the Trained Agents} For training, the agent is immersed in a 2D scenario (Figure \ref{fig:TrainingProcess}A), receiving its reward based on equation \eqref{eq:RewardFunction}, which is represented as an indicator function that assigns the reward to the agent only when it is inside the patch, and zero otherwise. This reward model is parameterized to the values of \(N\) and \(\lambda\) taken from \cite{wispinski2023adaptive}, set at \(N=30\) and \(\lambda = 0.01\). Thus, due to the nature of the reward function (\(r(n)\)), the agent faces a task of discounted patch foraging. The agent is in a 2D environment measuring $248\times456$ pixels and has a $64\times64$ pixel observation of the environment (note that for illustrative purposes, the observation in each image shown has a higher resolution). In each episode, a set of preloaded maps becomes available, and one of these maps is selected under a uniform probability distribution at the start of the episode. There are four maps differing from each other by the distance between patches (Figure \ref{fig:TrainingProcess}A), hence the probability of a map appearing in each episode is \(P(\text{map}) = \frac{1}{4}\). Each episode consists of 1500 steps, and during each training step when the agent is inside the patch, in addition to being assigned a reward, the agent receives a visual cue about the patch's state, which is the patch changing color in RGB-\(\alpha\) format. It starts as green \([255, 0, 0, 255]\) and darkens progressively with \(\frac{r(n)}{N}\), decreasing the \(\alpha\) value until it eventually becomes dark, i.e., as \(n \rightarrow \infty, \alpha \approx 0 \).

All the above describes the dynamics of the environment, which are the same for each of the algorithms used to tackle the task of discounted patch foraging: two model-free algorithms (R2D2, PPO) and one model-based algorithm (DreamerV2). The architectures of the three agents are constructed and undergo training for the $1.5 
\times 10^{6}$ steps (Figure \ref{fig:TrainingProcess}B). The learning curves shown in that figure, were constructed from three training runs, allowing us to generate an error graph with its respective mean (solid line) and standard deviation (shaded region). Specifically for DreamerV2, exponential curve smoothing was applied using a parameter $\omega=0.95$. Given a sequence of training values $\mathbf{X_T} = [x_1, x_2, \dots, x_n]$, where $x_i$ is the return value at a specific step, and $\omega \in [0, 1]$ is the weight of the parameter, the greater it is, the more smoothing occurs. This can be synthesized as follows: $s_i = \omega x_i + (1-\omega)x_{i-1}$, where $s_i$ is the smoothed value at time $i$.

%%%%%%%%%%%%%%%%%%%%%%%%
\subsubsection{R2D2 Architecture} The network utilized a learning rate of $1\times10^{-4}$ and a discount factor of 0.997. The training batch size was set at 1280, and the Adam optimizer was configured with an epsilon of 0.001, operating across two parallel workers. The training strategy employed complete episodes for batch processing, with a zero-initialized state setting. The network's architecture included convolutional filters with configurations of $\left(\left[\text{number of filters}, [\text{kernel Size}], \text{stride} \right]\right)$ [16, [8, 8], 8] and [128, [11, 11], 1], followed by RELU activation. This was complemented by fully connected hidden layers with 256 neurons, also using RELU activation. A LSTM network was integrated, featuring a cell size of 256 and utilizing previous actions, but not previous rewards.

%%%%%%%%%%%%%%%%%%%%%%%%
\subsubsection{PPO Architecture} For the PPO neural network model, we integrate advanced reinforcement learning techniques with deep learning structures. The configuration employs a critic baseline, utilizing the Generalized Advantage Estimator (GAE) with a lambda parameter of 1.0, and an initial coefficient for Kullback–Leibler (KL) divergence set at 0.2. Our model collects batches of 200 timesteps from each worker, with a training batch size of 4000 and an SGD minibatch size of 128. This setup is complemented by a stochastic gradient descent (SGD) learning rate of $5\times 10^{-5}$, over 30 iterations per outer loop. The model does not share layers for the value function, and the value function loss coefficient is set at 1.0. An entropy coefficient of 0.0 is used, along with a PPO clip parameter of 0.3 and a value function clip parameter of 10.0. Regarding the model's architecture, the convolutional filters and LSTM details are consistent with those used in the R2D2 model.

%%%%%%%%%%%%%%%%%%%%%%%%
\subsubsection{DreamerV2 Architecture} In our DreamerV2-based model implementation for complex environment interaction, we configured the world model to employ a dataset size of 2 million with a FIFO strategy and a batch size of 50, alongside sequence lengths of 50. The Recurrent State-Space Model (RSSM) \cite{Hafner2019} utilizes 600 units with discrete latent dimensions and classes set to 32, and the world model learning rate is set at $2 \times 10^{-4}$, employing a \texttt{tanh} reward transformation. The behavior policy relies on an imagination horizon of 15 steps with a discount factor of 0.995, using the Generalized Advantage Estimator with a lambda value of 0.95. Our actor and critic networks follow an actor-parameter mixing strategy with a learning rate of $1 \times 10^{-3}$ for the actor and $1 \times 10^{-4}$ for the critic, updating the slow critic every 100 steps. Common parameters across models include a policy steps per gradient step of 4, a Multilayer Perceptron (MLP) with 400 units across four layers, and a gradient clipping at 100. Additionally, we set the Adam optimizer with a learning rate of $1 \times 10^{-4}$, an epsilon of $1 \times 10^{-5}$, and a weight decay of $1 \times 10^{-6}$. The model's exploration and exploitation behaviors are fine-tuned to balance the intrinsic and extrinsic reward signals, with a significant focus on KL divergence for model learning optimization.

%%%%%%%%%%%%%%%%%%%%%%%%
\subsection{Probe Policy Methods} To evaluate the efficacy of the policies derived from the training phase, we utilized four distinct test scenarios: the first with \(\bar{x}_1 = 3\), the second with \(\bar{x}_2 = 5\), the third with \(\bar{x}_3 = 7\), and the fourth with \(\bar{x}_4 = 9\) (Figure \ref{fig:TrainingProcess}A). This testing phase involved conducting 25 individual experiments for each scenario, each comprising 1500 steps, thereby providing a statistical basis to assess the agent's performance in the patch foraging task. Our analysis focused on two primary metrics: the total number of steps taken by the agent, juxtaposed with the theoretical predictions of the MVT through the optimization of the function \(E_n: n \rightarrow \mathbb{R}\); and the final score achieved by the agent in each gameplay scenario. The former metric involved a direct comparison of the MVT's optimal step count against the mean number of steps recorded in the RL algorithm. The latter metric, the gameplay score, was determined by noting the score at the final step of each simulation. This dual-metric approach is instrumental in evaluating the agent's performance against both the MVT's theoretical optimum and its actual effectiveness within the game environment. It is noteworthy that the reported results excluded the R2D2 algorithm, owing to its demonstrated lack of learning during the training phase. To visually represent these findings, we plotted two quartiles around the median and extended to 1.5 times the interquartile range, thereby forming comprehensive boxplots for each tested policy. The policies subjected to this evaluation were PPO and DreamerV2.

%%%%%%%%%%%%%%%%%%%%%%%%
\subsubsection{Agent Trajectories} To draw the agents' trajectories, we recorded the agent's position at each simulation step in such a way that the information was available at the end of each experiment. We conducted 25 experiments of 1000 steps each, for every available test scenario ($\bar{x}_1 = 3, \bar{x}_2 = 5, \bar{x}_3 = 7, \bar{x}_4 = 9$ spaces), for both DreamerV2 and PPO. After executing each experiment for each environment, we concatenated the agent's position from each experiment to create an array of 25000 positions per scenario. This data was then used to draw the map that represents the positions where the agent was located, thereby generating the trajectories taken when testing the policy.
\subsection{Dreams Analysis}
The analysis of the agent's dreams begins with the Actor-Critic architecture of the DreamerV2 algorithm. The same architecture used in training was modified slightly for analysis: in training, the agent \say{dreams} and predicts the sequence of recurrent states \(\hat{h}_2, \hat{h}_3, \dots, \hat{h}_H\), where \(H\) is the prediction horizon, latent representation spaces \(\hat{z}_2, \hat{z}_3, \dots, \hat{z}_H\), rewards \(\hat{r}_2, \hat{r}_3, \dots, \hat{r}_H\), actions \(\hat{a}_2, \hat{a}_3, \dots, \hat{a}_H\), and state values \(\hat{v}_2, \hat{v}_3, \dots, \hat{v}_H\), starting from an observation \(x_1\), which is compressed into a latent representation space \(z_1\) and a recurrent state \(h_1\). For testing, we set \(H=50\) and reconstructed the predicted latent representation space \(\hat{z}_{2:H}\) using a decoder to visualize the agent's \say{dreams} (Figure \ref{fig:DreamAnalysis}).

Of course, the other elements of this \say{imagined} POMDP are used to later form the t-SNE projection, which for the obtained results had a perplexity parameter of 30. In the process, \(20 \times 10^{3}\) latent representation spaces of $32 \times 32$ dimensionality were taken and flattened into a row vector of 1024 positions, forming an array of \(20 \times 10^{3}\) rows by 1024 columns. This array fed into the t-SNE algorithm maps from this high-dimensional array to one of \(20 \times 10^{3}\) rows by 2 columns. The color of each cluster depends on the state value, which can be interpreted as a scalar number indicating how beneficial it is for the agent to be in a given state in terms of the expected sum of rewards (Figure \ref{fig:TSNE}).

%%%%%%%%%%%%%%%%%%%%%%%%%%%%%%%%%%%%%%%%%%%%%%%%%%
\subsection{Results}

\begin{figure*}[ht!]
    \centering
    \includegraphics[width=0.8\textwidth]{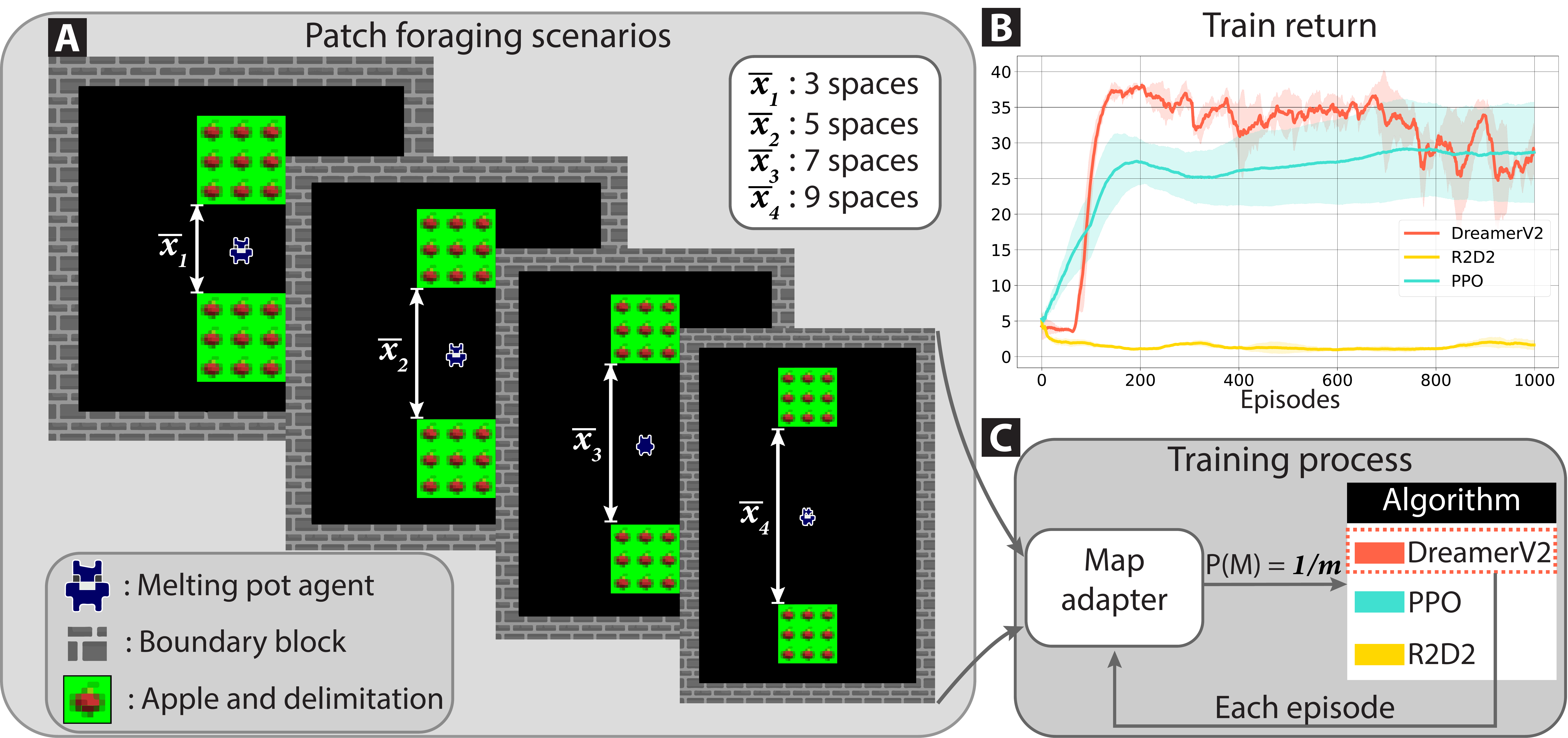}
    \caption{Training Process. (A) Scenarios generated in Melting Pot for training reinforcement learning algorithms, where $\bar{x}$ progressively increases in each map ($\bar{x}_4 > \bar{x}_3 > \bar{x}_2 > \bar{x}_1$), ensuring diverse maps during training. (B) Training returns for each algorithm: two model-free (PPO and R2D2) and one model-based (Dreamer V2). Each curve is calculated based on the average of three different training runs. For DreamerV2, an exponentially smoothed curve is shown (with a smooth coefficient of $\omega=0.95$). We track the train return behavior of each of the tested algorithms (y-axis) over the learning trajectory (x-axis in episodes). We plot the average trajectory per algorithm (and a standard deviation associated with the mean). (C) The training process involves an adapter that presents different maps to the agent in each episode, based on a uniform probability distribution $P(\text{Map}) = \frac{1}{m}$, where $m = 4$ is the number of maps.}
    \label{fig:TrainingProcess}
\end{figure*}

Our study examines two categories of reinforcement learning algorithms to test the hypothesis that world model structures enable agents to achieve efficient behaviors aligned with the MVT: model-based and model-free strategies (Figure \ref{fig:TrainingProcess}). We have selected DreamerV2, an algorithm that directly learns a compact state representation from high-dimensional image inputs to predict future outcomes during the learning process, as the representative of model-based RL algorithms. We have compared this algorithm against two model-free algorithms with differing approaches to policy learning: Proximal Policy Optimization (PPO), which is an on-policy model-free algorithm, and Recurrent Replay Distributed DQN (R2D2), an off-policy model-free algorithm \cite{schulman2017proximal,kapturowski2018recurrent}. On-policy learning involves policy refinement using the same policy currently being optimized, whereas off-policy learning leverages past experiences regardless of the current policy. This comparison aims to contrast methods with and without a model during the learning process and explore variations in policy learning methodologies, both on- and off-policy.

%%%%%%%%%%%%%%%%%%%%%%%%%%%%%%%%%%%%%%%%%%%%%%%%%%
\subsubsection{Learning Dynamics in Reinforcement Learning.}
As shown in Figure \ref{fig:TrainingProcess}B, the return per episode during the agent's training is illustrated. Notably, DreamerV2 demonstrates accelerated learning, as it outperforms the other two algorithms over 200 episodes, showing a more rapid growth compared to the model-free algorithms PPO and R2D2. It is important to clarify that none of these models underwent hyperparameter tuning, meaning the results presented in this section for both the training process and the testing of emerging policies were achieved without hyperparameter optimization. 

During the training process, the policy with the highest return is stored. That is, from the historical episodes, the policy that achieves the highest return for that episode is retained. This policy will then be evaluated against the scenarios used to train the agents, as shown in Figure \ref{fig:TrainingProcess}A.
\begin{figure*}[ht]
    \centering
    \includegraphics[width=0.8\textwidth]{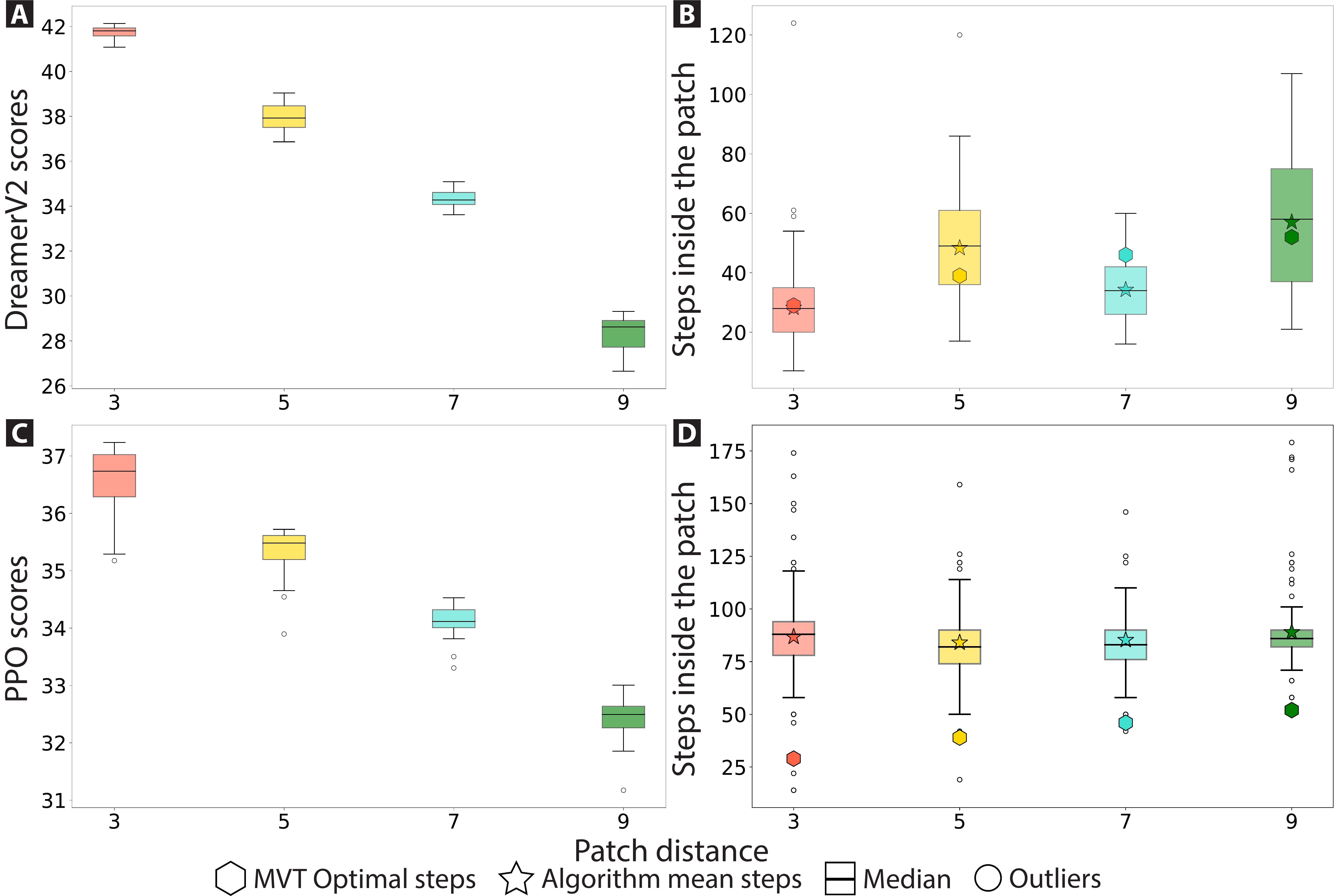}
        \caption{Performance of Reinforcement Learning Algorithms. (A) Score achieved by DreamerV2 at the final step of a simulated episode where the policy was tested for each proposed scenario, that is, when \(\bar{x}_4 = 9, \bar{x}_3 = 7, \bar{x}_2 = 5, \bar{x}_1 = 3\). The median score from the set of experiments is plotted (\(\pm\) one quartile), with each color representing a different test scenario. (B) Comparison of DreamerV2's performance with the predictions of the Marginal Value Rule (hexagon), showing the mean (star) and median (\(\pm\) one quartile) of the experiment set. (C) Under the same conditions, the score achieved by PPO on the same test scenarios is displayed. (D) Comparison between PPO and the Marginal Value Theorem, under the same conditions.}
    \label{fig:MainResults}
\end{figure*}
If we closely examine the performance of R2D2, it lacks substantive meaning for policy analysis, as the algorithm for this task shows no signs of learning. R2D2 may have struggled in our patch-foraging task because its off-policy, value-based learning relies on replayed experience and long-horizon credit assignment, which can make it less sensitive to the gradual decline in reward that defines patch depletion. Without an explicit predictive model, the agent may overestimate the value of staying in a patch and fail to adjust its behavior quickly enough to approximate marginal-value–like decisions. Because it performed poorly compared to the other two trained models, PPO and DreamerV2. Consequently, from this point forward, our discussion will focus only on the results obtained from these algorithms, with special attention on DreamerV2.

Based on Commons Harvest scenario of Melting Pot, we created a patch-foraging environment in this suit, as shown in Figure \ref{fig:EnvironmentDynamics}B. We have found it beneficial for the learning process to delineate the patch with a green background, as seen in Figure \ref{fig:TrainingProcess}A. This modification simplifies the determination of whether the agent is within the patch and provides visual cues to the agent about the patch boundaries. Another immediate visual clue available to the agent is the reward, indicated by the patch's color change in proportion to \(\frac{r(n)}{N}\).

%%%%%%%%%%%%%%%%%%%%%%%%%%
\subsubsection{Strategy Validation: Implementing and Assessing Optimal Policies.}
We present the results obtained from testing the policies developed during training. Initially, let us focus on Figure \ref{fig:MainResults}A, which describes the score achieved by DreamerV2 when testing the best-performing policy from training across established test scenarios. These scenarios are designed to manipulate the distance between patches, thereby indirectly altering the travel time required for agents to move from one patch to another. This variation also affects the richness of resources in the environment, similar to many experiments on adaptive food-search behavior in patchy environments observed in animals \cite{COWIE1977, Kacelnik1984}. In such settings (where distance is altered), our agent's behavior is at least desirable, as increasing the distance between patches results in lower scores, indicating that the spatial arrangement of patches plays a crucial role in the agent's decision to spend more time in a patch. This models an understanding of the scenario dynamics, as the agent learns to associate the cost of switching patches with increasing distance. Notably, the box plots highlight minimal variability in scores, where similar scores were consistently achieved across experiments.

\begin{figure*}[ht]
    \centering
    \includegraphics[width=0.8\textwidth]{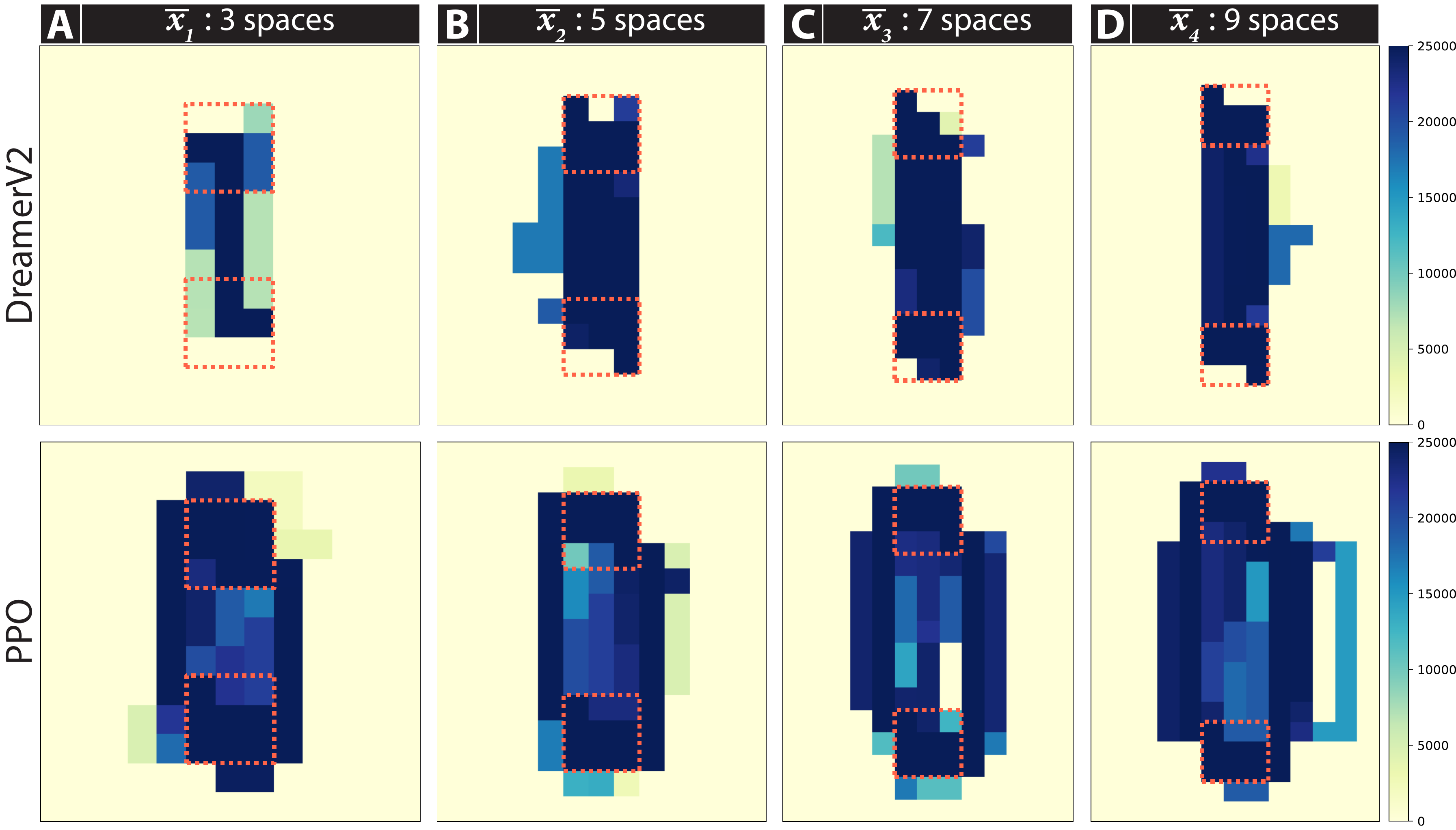}
    \caption{The trajectories followed by the trained agents are illustrated in the figures. The top row displays the paths taken by the agent trained with DreamerV2, while the bottom row shows the paths of the agent trained with PPO. Columns (A), (B), (C), and (D) represent different agent trajectories when the patches were spaced 3, 5, 7, and 9 units apart, respectively. The dotted lines in all the figures indicate the patch boundaries, and the color of each tile reflects the number of steps the agent took on that particular tile.}
    \label{fig:AgentsPaths}
\end{figure*}

Compared with MVT, Figure \ref{fig:MainResults}B displays the step count of the agent trained with DreamerV2, which exhibits quite interesting behavior in terms of the number of steps taken in each of the proposed scenarios. It is evident that, on average, the DreamerV2-trained agent takes a number of steps comparable to what the MVT dictates for each scenario. In at least three out of the four scenarios, the optimal value of the MVT falls within the $\pm$ 1 quartile of the distribution yielded by the set of experiments. In the scenario where this is not met, at least the value indicated by the MVT is within the interquartile range of the diagram, which would be the worst-case scenario. Analyzing these diagrams from the perspective of behavioral ecology theory also reveals interesting findings, such as the tendency, in at least three of the four scenarios, for agents to exceed the number of steps \say{expected} to be taken on the patch. This is consistent with biological animal behavior \cite{Nonacs2001, Carter2016}.

So far, we have shown and described results obtained using model-based RL algorithms. However, in this patch foraging task, it is worthwhile to test at least one model-free algorithm. Although PPO shows an increasing trend in Figure \ref{fig:TrainingProcess}B, it is important to examine how its policy performs across the test scenarios. The performance of PPO in terms of scores is also consistent, in that it achieves lower scores as the distance between patches increases. However, there is a noticeable difference in behavior here: if we look closely at Figure \ref{fig:MainResults}C, the scores at the end of each episode show slightly more variability compared to those obtained with DreamerV2, and outliers appear due to this variability in the scores. This indicates that in terms of this metric, DreamerV2 achieves better scores for each scenario and the results are less variable.

Regarding the metric that measures how closely the performance of the reinforcement learning algorithm aligns with the optimal value of the MVT, PPO shows a performance that is significantly divergent when the MVT is used as a baseline. Although it is true that biological agents in some scenarios tend to spend more time in patches, the difference is large enough that the MVT value does not even fall within the interquartile range, highlighting a clear disadvantage of the model-free algorithm in approximating the optimal behavior dictated by the MVT. In summary, when testing the emergent policy of PPO post-training, this algorithm fails to approach the standards set by the MVT. Another evident factor is the significant number of outliers present in the experimental results, indicating considerably more outliers than those found in the model-based algorithm, which reflects the high variability illustrated in Figure \ref{fig:MainResults}D. In none of the four test scenarios does the metric thrive; notably, in the scenario where the agent should ideally take the fewest steps, it actually takes the most on average, and this is where the most distant outliers from the experimental set are found.

An interesting aspect of the policies resulting from the agents' training is to review the type of actions carried out by the agent in each of the available scenarios for evaluating such policy. The results show a clear difference in the behaviors of each agent (Figure \ref{fig:AgentsPaths}). For DreamerV2, the agent's trajectories indicate an efficient performance, particularly in the scenario with $\bar{x}_1 = 3$ spaces, where the agent almost always follows the same route (Figure \ref{fig:AgentsPaths}A). Note that in the figure, the dark blue color corresponds to a frequently traveled route, while the moderately yellow color indicates less traveled routes. This justifies the previous result observed in Figure \ref{fig:MainResults}B, especially since in this scenario DreamerV2 is on average closest to the MVT. Analyzing the other scenarios, DreamerV2 continues to demonstrate clear efficiency, generally keeping all its trajectories confined within the middle of the map. This indicates that the agent frequently uses paths that lead directly to the other patch, avoiding wasting energy on longer routes, thereby showing a clear efficiency and understanding of the scenario it is immersed in. On the other hand, the counterpart that does not use world models (PPO), exhibits considerable energy wastage in its movements, taking longer routes to reach the patch and performing actions that cause the agent to meander, highlighting the inherent weakness of model-free algorithms. Furthermore, as the distance increases, the agent takes even longer routes, exhibiting this wandering behavior more markedly.

To conclude, the effectiveness of the model-based algorithm may be attributed to the nature of creating a model. Specifically for DreamerV2, a world model is hypothesized to encapsulate fundamental concepts that bring about quasi-optimality in the patch foraging problem, in comparison to the MVT. This should demonstrate some congruence with evidence about animal behavior and how having certain knowledge of the world allows for the anticipation of situations, enabling planning to avoid unfavorable scenarios and, thereby, developing strategies that are evolutionarily stable over time.

\section{Discussion}
This work suggests that model-based reinforcement learning algorithms, especially those leveraging world models, potentially exhibit performance comparable to that outlined by the Marginal Value Theorem. However, the findings reveal a tendency for the artificial agent to remain in the patch longer than it should. One assumption that might bias the MVT relates to the agent having complete information about the environment it inhabits. This, of course, does not occur in nature, and evidence suggests that in patch foraging tasks, animals tend to exhibit self-controlled behavior as opposed to the impulsive decision to leave the patch when rewarded instantaneously \cite{Hayden2016}. In this respect, the findings are in line with behavioral ecology states should happen (Figure \ref{fig:MainResults}). This alignment may be due to how the environment setup takes a partial observation of the environment, as seen in Figure \ref{fig:EnvironmentDynamics}B. The agent also does not have complete information about the environment; nevertheless, the results show that there is a degree of uncertainty associated with the decisions made by the agent under the learned policy. This is because actions from the learned policy are drawn from a probability distribution, and the agent must sample this distribution to take the action that maximizes the reward, thus introducing some uncertainty and variability in the policy evaluation results. Another factor contributing to this uncertainty is the way apples appear on the patch, as their reappearance or absence depends on the number of neighbors an apple has and a random component.

%%%%%%%%%%%%%%%%%%%%%%%%%%%%%%%%%%%
\subsection{The Role of World Models in the Emergence of Optimality.}In animal behaviors, particularly in the contexts of searching and foraging for food, there appears to be a dual influence: one from the direct observations of their surrounding environment, and the other from a mentally conceptualized world model. This world model acts as a cognitive framework, enabling animals to make predictive judgments about potential occurrences. Possessing an understanding of this world model confers an ecological advantage, providing the animal with the capacity to base decisions not merely on immediate environmental cues but also on anticipated future scenarios \cite{Leslie1991,Coulter2023}. This concept of a world model as an evolutionary benefit in animals serves as a biological muse for the development of learning models. In the realm of artificial intelligence, world models are increasingly seen as instrumental in engendering optimal behaviors in foraging contexts. These models are distinctively designed to facilitate policy learning indirectly, through simulated experiences before applying the learned strategies in real-world scenarios \cite{Ha2018}. LeCun's proposition of an autonomous intelligence model includes a dedicated module for constructing the world model representation. This module empowers the agent with the capability to forecast potential future states of the world, contingent on sequences of hypothesized actions \cite{lecun2022path}.

\begin{figure*}[ht!]
    \centering
    \includegraphics[width=0.85\textwidth]{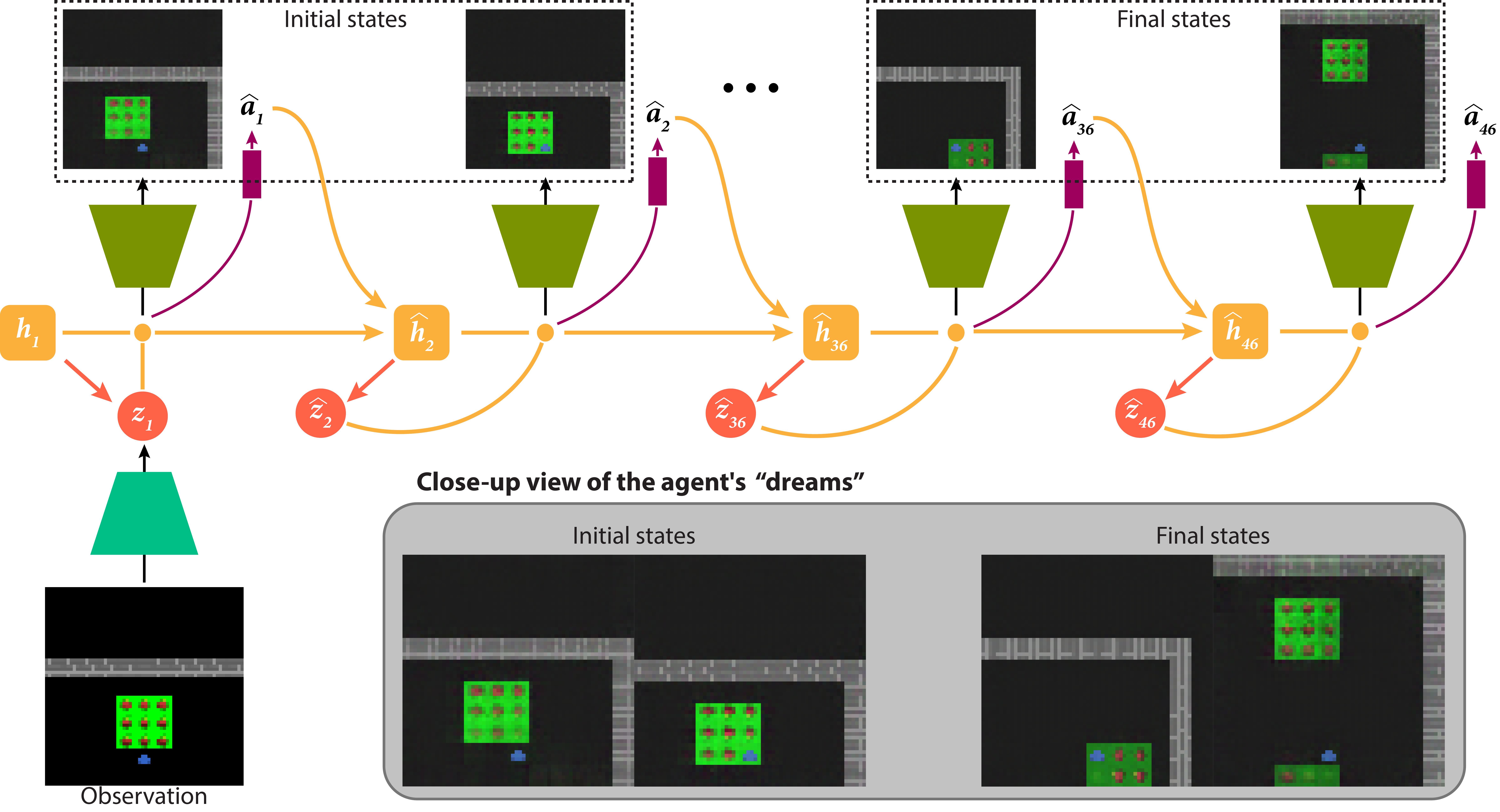}
        \caption{Sampled initial and final states from predicted trajectories. Utilizing the actor-critic structure of DreamerV2 and the world model assimilated by the agent, this figure illustrates trajectories envisioned by the agent under a partial observation scenario, which is essentially, an image. The Recurrent State-Space Model operates through a sequence of deterministic recurrent states \( h_t \), computing at each step a posterior stochastic state \( z_t \) that integrates information from the current image. Beginning with a specific state ($h_1$) and observation, the agent envisions potential trajectories. These can be reconstructed through a decoder, which reveals these envisioned scenarios.}
    \label{fig:DreamAnalysis}
\end{figure*}

The observed effectiveness in model-based algorithms, particularly in DreamerV2, underscores the critical role of world models in the emergence of optimal foraging behavior. These models, by capturing and synthesizing complex environmental information, enable agents to make more informed and adaptive decisions. To delve deeper into the study of what world models capture in the trained agent, it is necessary to utilize the actor-critic model that the agent possesses (Figure \ref{fig:DreamAnalysis}). The premise is that there is a learned representation of recurrent states \(h_t\) along with the dynamics of the world, which is precisely referred to as the world model—a compilation of experiences gathered by the agent during the initial stages of training, subsequently generating a policy that is refined as the representation of that world model becomes more accurate. We have adapted this architecture to enable a visual interpretation of what happens when the agent predicts future events, known as \say{dreams.} It should be noted that this adaptation was made only during the policy testing process.

A revealing aspect demonstrated by the agent's \say{dreams} is the accuracy with which it can replicate the dynamics of the world while dreaming. Figure \ref{fig:DreamAnalysis} displays the initial and final states of a sample dream that the agent experiences, starting from a recurrent state \(h_1\) and an observation embedded in a latent representation space \(z_1\). Based on these elements, an estimation of the subsequent sequence of recurrent states and latent representation spaces is generated, all products of the agent's imagination. The agent is capable of imagining both the inherent dynamics of the patch, such as the color change (representing the resource richness of the patch) and the amount of energy it can extract from it based on exponentially discounted rewards, as well as the spatial configuration of the foraging scenario (as can also be seen in Figure \ref{fig:AgentsPaths}). This reveals the agent's understanding and adaptability, particularly when the policy learned during training is put to the test.

We attribute the adaptability of an agent trained with RL entirely to the concept of a world model. Model-free algorithms pay the price associated with their direct interaction with the world and the updating of the value associated with that state through the process of trial and error in actions within the environment. Visualizing the agent's behavior is crucial when evaluating the efficiency of the agent in the foraging task. Mapping the trajectories followed by the agent is useful for identifying the type of behavior exhibited. In the case of model-free algorithms, there is a wandering behavior as seen with PPO (Figure \ref{fig:AgentsPaths}) \cite{wispinski2023adaptive}, where the agent is far from performing actions that lead to maximizing equation \eqref{eq:EnergyIntakeAdapted}. In this regard, having a world model allowed our DreamerV2 agent to invariantly model the distance between patches during the learning process, which led not only to behavior close to the optimum prescribed by the MVT (Figure \ref{fig:MainResults}B) but also to more efficient routes between patches.

%%%%%%%%%%%%%%%%%%%%%%%%%%%%%%%%%%%
\subsection{Parallels Between Biological Agents' Planning Abilities and Predictive Behavior in Artificial Agents.}
A fascinating aspect of model-based reinforcement learning is its parallel to the planning and predictive behavior observed in many biological agents. The ability to anticipate and plan for the future, traditionally seen as a hallmark of biological intelligence, appears to also emerge in artificial agents trained with advanced reinforcement learning approaches. However, the hypothesis that animals can or cannot plan for the future has been a topic of discussion within the behavioral ecology and neuroscience communities, as there is still debate over whether non-human individuals are capable of planning for the future. The Bischof-Köhler hypothesis implicitly suggests the idea that non-human individuals may be incapable of mental time travel, rendering them unable to dissociate a past or future state from their current one. A fully satiated animal, for instance, might not be able to comprehend that it will feel hunger in the future \cite{Suddendorf1997}.

\begin{figure}[ht!]
    \centering
    \includegraphics[width=0.6\textwidth]{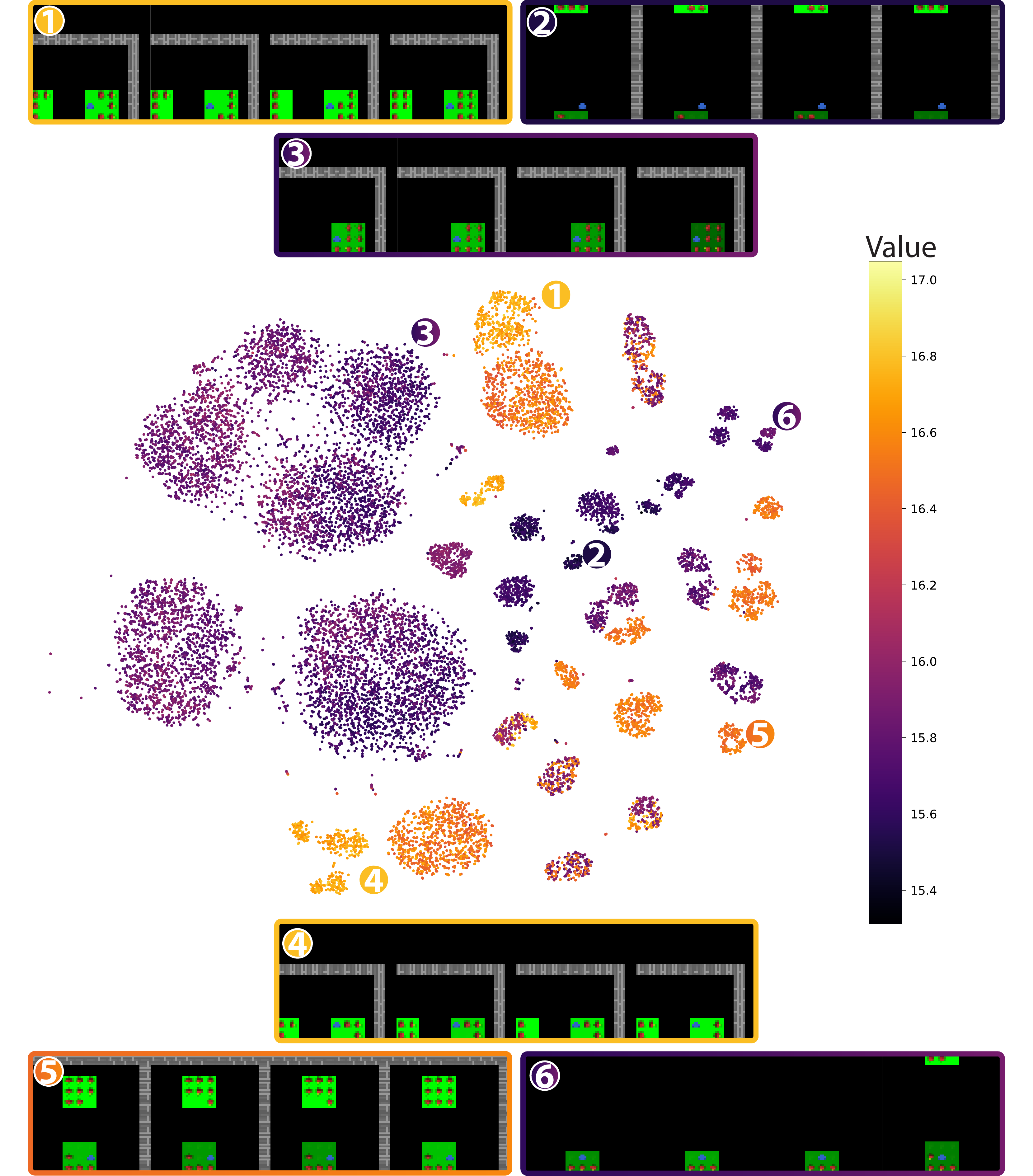}
    \caption{Insights derived by the DreamerV2-trained agent in optimizing equation \eqref{eq:EnergyIntakeAdapted}. The agent demonstrates the ability to discern key notions such as the patch spatial arrangement and its color transitions. Featured here is a t-SNE projection depicting various latent states, along with the sequence of observations utilized to deduce these states. Each state is distinguished by a color correlating with its state value, showcasing that these concepts lead to a distinct segregation of the states. This suggests that the established policy create a coherent and natural segmentation within the state-space. The numbers represent areas of interest for study, with each of these areas exhibiting distinct and separate concepts.}
    \label{fig:TSNE}
\end{figure}

Future-oriented planning refers to behavior directed towards future outcomes and shaped by learning. For this reason, it is at least understandable why one would want to analyze elements of future planning in learning agents, as learning and behavior oriented towards the future can be studied from the perspective of artificial intelligence, which in this case is capable of handling the task of foraging. There are at least three elements that constitute future planning \cite{Clayton2003}:

\begin{enumerate}
    \item \textbf{Content}: Guessing what might happen, and where and when it will happen, based on past experiences.
    \item \textbf{Structure}: Assembling a plan that addresses the what, where, and when.
    \item \textbf{Flexibility}: Utilizing memory and knowledge in a manner that can be altered and adapted as needed for planning the future.
\end{enumerate}

We have identified these three elements during the analysis of the agent's dreams. In fact, they are presented in such a way that they can be directly associated. For instance, predicting what will happen, where, and when is directly linked to the agent's \say{dreams.} Within these dreams, it is observable that the agent anticipates the depletion of resources on the patch, and can discern the moment to leave the patch due to the scarcity of resources (\textit{content}). In this context, what will happen equates to changing patches, where refers to the spatial location of the agent within the simulation environment, and when is contingent upon the resource availability of the patch, occurring when resources reach a level prompting the agent to decide to depart. The \textit{structure} then emerges from this world of dreams, forming a plan that answers what, where, and when, and this plan is precisely the structure illustrated in Figure \ref{fig:DreamAnalysis}. The \textit{flexibility} in this scenario relates to the agent’s adaptability in developing a policy that adjusts to changes in the spatiality of the patches, that is, the distance between patches.

Figure \ref{fig:TSNE} is a testament to the learning process behind future planning, as the agent can represent concepts such as distance or resource scarcity in a patch. This projection of the latent states encoded by the agent shows that the states with higher value are associated with the amount of resources available in the patch, and the less valuable states occur when the agent does not perceive the other patch within its observation. This demonstrates a high level of learning, showcasing the agent's ability to model the dynamics of the world it is immersed in. A key point we wish to highlight is the success of this task-specific foraging through the use of this discrete representation space that encapsulates world dynamics, as this is crucial in the emergence of intelligence \cite{ma2022principles}. Furthermore, it also enables generalization in the face of novel scenarios \cite{Gomez2023}.

\section{Conclusion}
In this work, we demonstrated that model-based reinforcement learning agents equipped with learned world models naturally develop patch-foraging behaviors aligned with the Marginal Value Theorem. The contribution of this study is therefore conceptual rather than algorithmic: instead of proposing a new method, we utilized a controlled set of environments and comparative analyses to show that anticipatory representations enable existing agents to approximate optimal patch-leaving decisions and exhibit behavioral patterns comparable to those observed in biological foragers. These findings highlight the importance of predictive internal models in supporting adaptive, interpretable decision-making and show how principles from ecological optimality can potentially inform the development of more robust AI systems.

Future studies in this type of scenario involve extending this research to a multi-agent foraging environment and use this work as a foundation to analyze community behavior in agents. This prospect is particularly appealing and poses a challenge, as the emergence of group behaviors in artificial intelligence entails a higher degree of complexity. Consequently, a future endeavor would be to implement a foraging scenario with two or more agents to observe the emergence of behaviors comparable to social foraging theories. Here, roles within the agent society could be examined. Using elements like the MVT, which is integral to social foraging theory, we can create indicators of agent optimality \cite{Caraco2000}. Studying these indicators would allow us to explore the emergence of complex behaviors present in a biological foraging society.

\bibliographystyle{spp-bst}
\bibliography{bibfile}
\end{document}